\documentclass{svproc}
\usepackage{url}

\usepackage[T1]{fontenc}
\usepackage{graphicx, verbatim}

\usepackage{epsfig}
\usepackage{amsmath}
\usepackage{mathtools}
\usepackage{amsfonts}       

\usepackage[nameinlink,capitalize]{cleveref}
\usepackage{booktabs}       
\usepackage{subfigure}
\usepackage{algorithm}
\usepackage{algpseudocode}
\usepackage{soul}
\usepackage{ragged2e}
\usepackage[usenames,dvipsnames]{xcolor}
\usepackage{microtype}

\begin{document}
\mainmatter              
\title{XSRD-Net: EXplainable Stroke Relapse Detection}
%
%
\author{Christian Gapp\textsuperscript{\textbf{*}}\inst{1,2}
	\and
	Elias Tappeiner\inst{1}
	\and
	Martin Welk\inst{1}
	\and 
	Karl Fritscher\inst{2}
	\and
	Stephanie Mangesius\inst{3,5}
	\and
	Constantin Eisenschink\inst{3}
	\and
	Philipp Deisl\inst{2,3,5}
	\and
	Michael Knoflach\inst{2,4}
	\and
	Astrid E. Grams \inst{3,5}
	\and
	Elke R. Gizewski\inst{3,5}
	\and
	Rainer Schubert\inst{1}
}
\authorrunning{C. Gapp et al.} 
%

\institute{Institute of Biomedical Image Analysis\\
	UMIT TIROL -- Private University for Health Sciences and Health Technology,	Eduard-Walln\"ofer-Zentrum 1, 6060 Hall in Tirol, Austria \and 
	VASCage -- Centre on Clinical Stroke Research, 6020 Innsbruck, Austria \and
	Department of Radiology, \and Department of Neurology, \and Neuroimaging Research Core Facility, Medical University of Innsbruck, 6020 Innsbruck, Austria\\
	\email{\{christian.gapp\textsuperscript{\textbf{*}}, elias.tappeiner, martin.welk, rainer.schubert\}@umit-tirol.at,\\ \{christian.gapp, karl.fritscher\}@vascage.at\\ \{stephanie.mangesius, constantin.eisenschink, philipp.deisl, michael.knoflach, astrid.grams, elke.gizewski\}@i-med.ac.at}
}

\maketitle              

\begin{abstract}
	Stroke is the second most frequent cause of death world wide with an annual mortality of around 5.5 million. Recurrence rates of stroke are between 5 and 25\% in the first year. As mortality rates for relapses are extraordinarily high (40\%) it is of utmost importance to reduce the recurrence rates. We address this issue by detecting patients at risk of stroke recurrence at an early stage in order to enable appropriate therapy planning.
	To this end we collected 3D intracranial CTA image data and recorded concomitant heart diseases, the age and the gender of stroke patients between 2010 and 2024.
	We trained single- and multimodal deep learning based neural networks for binary relapse detection (Task~1) and for relapse free survival (RFS) time prediction together with a subsequent classification (Task~2).
	The separation of relapse from non-relapse patients (Task~1) could be solved with tabular data (AUC on test dataset: 0.84).
	However, for the main task, the regression (Task~2), our multimodal XSRD-net processed the modalities vision:tabular with 0.68:0.32 according to modality contribution measures. The c-index with respect to relapses for the multimodal model reached 0.68, and the AUC is 0.71 for the test dataset.
	Final, deeper interpretability analysis results could highlight a link between both heart diseases (tabular) and carotid arteries (vision) for the detection of relapses and the prediction of the RFS time.
	This is a central outcome that we strive to strengthen with ongoing data collection and model retraining.
	
	\keywords{Stroke Relapse Prevention, Multimodal Fusion, RFS Time Prediction, Modality Contribution, Explainable AI}
\end{abstract}
\section{Introduction}
Stroke is the second most frequent cause of death world wide. The annual mortality is approximately 5.5 million \cite{Donkor2018-zg}. Stroke patients have a recurrence rate of 5--25\% in 1 year and 20--40\% in 5 years \cite{Caplan_2016}. Recurrences lead to death in nearly 40\% of cases that occurred within the first two years according to the North Dublin Population Stroke Study \cite{Callaly2016-dq}. Hence the probability for an acute stroke patient to suffer a relapse followed by death is alarming. Therefore, it is of utmost interest to detect patients that potentially suffer a relapse after initial stroke in order to enable proper treatment at an early stage.
This analysis focuses on arteriosclerotic and thrombotic strokes. Other etiologies, such as arterial dissections, are characterized by distinct pathophysiological mechanisms, and are comparatively less common.

Risk factors for stroke are widely researched and well known (high blood pressure, diabetes mellitus, high blood cholesterol etc.) \cite{Murphy2020-rf}. However, risk factors for relapses are yet rarely explored, as for example in \cite{Uzuner2023-gz}. Especially reasons for early relapses (occurred within 3 months) are even less known.
Therefore, with our work we want to search for new, significant biomarkers for relapses.

From April 2023 onwards we collected clinical and imaging data from patients with at least one ischemic cerebral event (ICE) in the time between 2010 and 2024. 
For each patient we have taken a 3D intracranial CTA image at the time of the first event. Furthermore, information about concomitant heart diseases, viz. CHD (coronary heart disease) and PAD (peripheral artery disease), has been recorded. The attributes age and gender complete the tabular data.
Finally, we could use data from 119 patients, including 78 non-relapse and 41 relapse cases.

\begin{figure}[t!]
	\centering
	\includegraphics[width=0.75\textwidth]{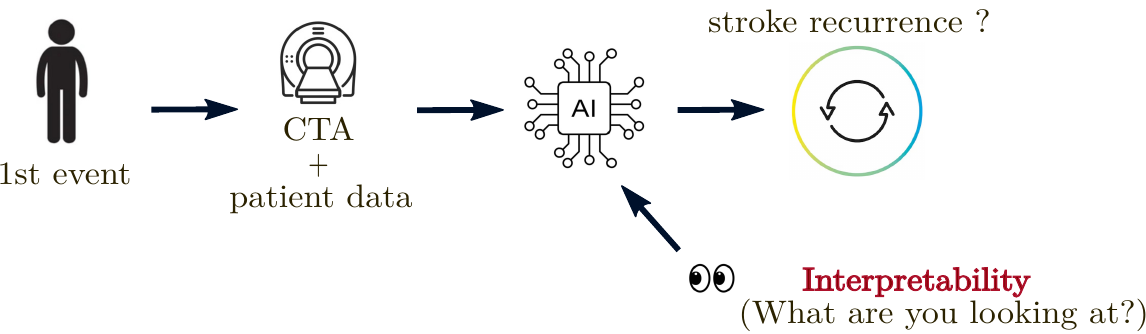}
	\caption{Schematic pipeline: Interpretable stroke recurrence detection.}
	\label{fig:schematic_pipeline}
\end{figure}

With deep learning based methods we train single-modal, viz. vision-only and tabular-only, and multimodal neural networks for binary classification (Task~1) into non-relapse and relapse cases and for the prediction of relapse free survival (RFS) times (Task~2), i.e., a regression problem. Subsequently a classification on the regression task is done additionally.
To better understand the role of each modality in solving the multimodal task, we apply the approach introduced in \cite{Gapp_MCI}. This analysis not only highlights the contribution of each data source but also uncovers novel insights into the underlying causes of stroke relapses. The overall methodology is illustrated in the pipeline shown in \cref{fig:schematic_pipeline}.

\section{Related Work}\label{sec:relatedWork}

In a population wide cohort study in Australia and New Zealand from
2008--2017 \cite{Peng2022-zj} authors
found that female sex and increasing age were associated with increased risk of mortality and recurrent stroke. Furthermore, death and recurrence of stroke were common after an acute stroke. The authors claim to focus research on secondary stroke prevention.

The recurrence rate after a stroke is already 5--25\% after just one year \cite{Caplan_2016}. According to \cite{Callaly2016-dq}, recurrences that occurred within the first two years after the initial event lead to death in almost 40\%. This numbers emphasize the importance of relapse prevention.

In \cite{StrokeOutcomeMultimodal23} a multimodal neural network was trained to predict a 90-day modified Rankin Scale (mRS) score for patients with acute ischemic stroke using MRIs and clinical data.
However, recurrences of stroke were not analyzed. 

A machine learning based stroke recurrence prediction was performed in \cite{Ding2024} using EHR (Electronic Health Record) data. However, image data was not considered. Results are useful for further epidemiological research.

In \cite{MultimodalStrokeModel2024} single-modal and multimodal neural networks were trained to detect ischemic stroke (IS) relapses using MRIs and clinical data. Top three negatively correlated risk factors for IS recurrence are total cholesterol, HDL (high-density lipoprotein), and diabetes mellitus. Top three positively correlated risk factors are LDL (low-density lipoprotein), smoking and history of heart disease. However, sex differences and the pathological relationship between diabetes and IS recurrence remains controversial as studies \cite{Chung2023-nt,Wang2023-ti}
conclude differently.

Authors from \cite{Uzuner2023-gz} claim that congestive heart disease and diabetes mellitus have a link to higher rate of multiple recurrences for ischemic stroke. 

In summary, to the best of our knowledge one can say that the presented approach for explainable prediction of stroke RFS time combined with relapse detection using intracranial 3D visual data together with patient-specific tabular data (in this work: age, gender and heart diseases) has not yet been explored. In particular, the integration of interpretability into this context is a novel contribution and provides valuable insights that have been largely unexplored in previous research.

\section{Multimodal Stroke Data}\label{sec:data}

\subsection{Data Generation}
Data was recorded as part of the project “Retrospective pilot project: Imaging Biomarkers For Vascular Diseases and Vascular Aging”. From April 2023 onwards we collected clinical as well as imaging data from patients with at least one ischemic cerebral event (ICE). Our patient collective comprized patients who were admitted to the Stroke Unit of the Department of Neurology from 2010 onwards.
Imaging data was recorded on the platform “syngo.share” (Version VA32C) by Siemens. Clinical data was recorded on a custom made database set up by an external company. Patients who received CT-angiographic imaging as part of their routine diagnostic work up were fully anonymized before further analysis.
In addition, the imaging data were labelled as to whether the patient had a cardiac or peripheral event, had a subsequent ICE, or had died.
The project was approved by the local institutional review board (IRB) of the Medical University of Innsbruck (EK-Nr: 1429/2021).

\subsection{Study Population}
The patients from the cohort are selected with respect to their RFS time. Every relapse patient must have a RFS time that is lower than every non-relapse patient's. 
We have set the upper RFS limit for relapse patients to $\kappa$\textsubscript{low} = 1,642 days. All relapse patients with a RFS < $\kappa$\textsubscript{low} are included, others unused. The lower RFS limit for non-relapse patients is $\kappa$\textsubscript{high} = 1,825 days. Only non-relapse patients with a RFS > $\kappa$\textsubscript{high} are processed. In addition, we have set a cut-off at 2,555 days for non-relapse cases.
Finally we can use data from 119 patients. 78 patients have only suffered one stroke so far, i.e., patients without a relapse (\emph{non-relapse}), and 41 have already had a relapse. For 23 of these, the RFS time is known. The other 18 relapse patients' RFS is unknown due to data lack. We do not need the RFS time for the binary classification. However, for the regression task we rely on the RFS. In order to use these 18 patients completely we estimated their RFS time during training with the predicted RFS time from the 23 other relapse patients (details see \cref{fig:rfs_estimation_progression}).

We have split the data into a train and a test dataset with a ratio of 80:20. The train dataset consists of 95 patients (including 32 relapses) and the test dataset has 24 patients (9 relapses).
%
Statistical details of the study population, summarizing demographic and clinical characteristics, are presented in \cref{tab:statistical_details_stuy_pop}.

\setlength{\tabcolsep}{10pt}
\begin{table}[h!]
 \centering
\caption{Baseline characteristics of the study population (N = 119).}
\begin{tabular}{llcc}
	\toprule
	characteristic & attribute & n & \% \\
	\cmidrule{1-4}
	\textbf{gender} & men  & 79 & 66.4 \\
	& women & 40 & 33.6 \\
	\addlinespace[2mm]
	\textbf{heart disease} & CHD only & 24 & 20.2 \\
	& PAD only & 20 & 16.8 \\
	& CHD + PAD & 6 & 5.0 \\
	& none & 69 & 58.0 \\
	\addlinespace[2mm]
	\textbf{relapse status} & no relapse & 78 & 65.6 \\
	& relapse & 41 & 34.4 \\
	\cmidrule{1-4}
	& & \multicolumn{2}{c}{mean $\pm$ SD} \\
	\cmidrule{3-4}
	\textbf{age (years)} & &\multicolumn{2}{c}{$69.10 \pm 10.18$} \\
	\bottomrule
\end{tabular}
\label{tab:statistical_details_stuy_pop}
\end{table}

\section{Methodologies}\label{sec:methods}

\subsection{Data Preprocessing}\label{para:handlingIncompleteData}
\paragraph{Image Registration}
We register all images to one fixed image in order to ensure similar properties for further data processing with deep learning methods. The fixed image is selected as the one that most closely reflects the typical shape among all images. The moving images are registered to the fixed image by applying affine transformations using the medical image registration library \emph{SimpleElastics}~\cite{SimpleElastics}.

\paragraph{Tabular Data Processing}
The attributes age, gender, CHD (coronary heart disease) and PAD (peripheral arterial disease) have to be preprocessed in order to be used for deep learning methods.
Age: 
Let $y_i$ be the age of a patient $i \in \{1, \dots, N\}$. Then the z-normalized age is
\begin{equation*}
	\hat{y}_i = \frac{y_i - \mu}{\sigma}, \quad \text{with} \quad
	\mu = \frac{1}{N} \sum_{i=1}^{N} y_i, \quad \sigma = \sqrt{\frac{1}{N-1} \sum_{i=1}^{N} (y_i - \mu)^2}.
\end{equation*}
Gender: For the gender we define
[0,1]: Female and [1,0]: Male.
Heart diseases:
CHD and
PAD are encoded as
$\lbrack 0,0\rbrack$: No disease, 
$\lbrack1,0\rbrack$: CHD, 
$\lbrack0,1\rbrack$: PAD, 
$\lbrack1,1\rbrack$: CHD + PAD.

\paragraph{Handling Incomplete Data}
We have some relapses with unknown RFS time due to missing second event times. However, we always know their maximum possible RFS time, which is determined by the time between the occurrence of the first event and a fixed stop date for the data collection.
Those with a maximum possible RFS~>~$\kappa$\textsubscript{low}, i.e., the upper limit for RFS times of patients with relapse, are removed. The others are used straight forward for Task~1: classification. However, for the regression task, viz. Task~2, their RFS time must be estimated during the trainings routine. For this we take the current predicted mean RFS time (at each training epoch) of all relapse patients with known ground truth RFS time. Through this dynamic update (see also \cref{fig:rfs_estimation_progression}), all patients can be fully integrated into the training routine.

\subsection{Architectures}

\begin{figure}[t!]
	\centering
	\includegraphics[width=\textwidth]{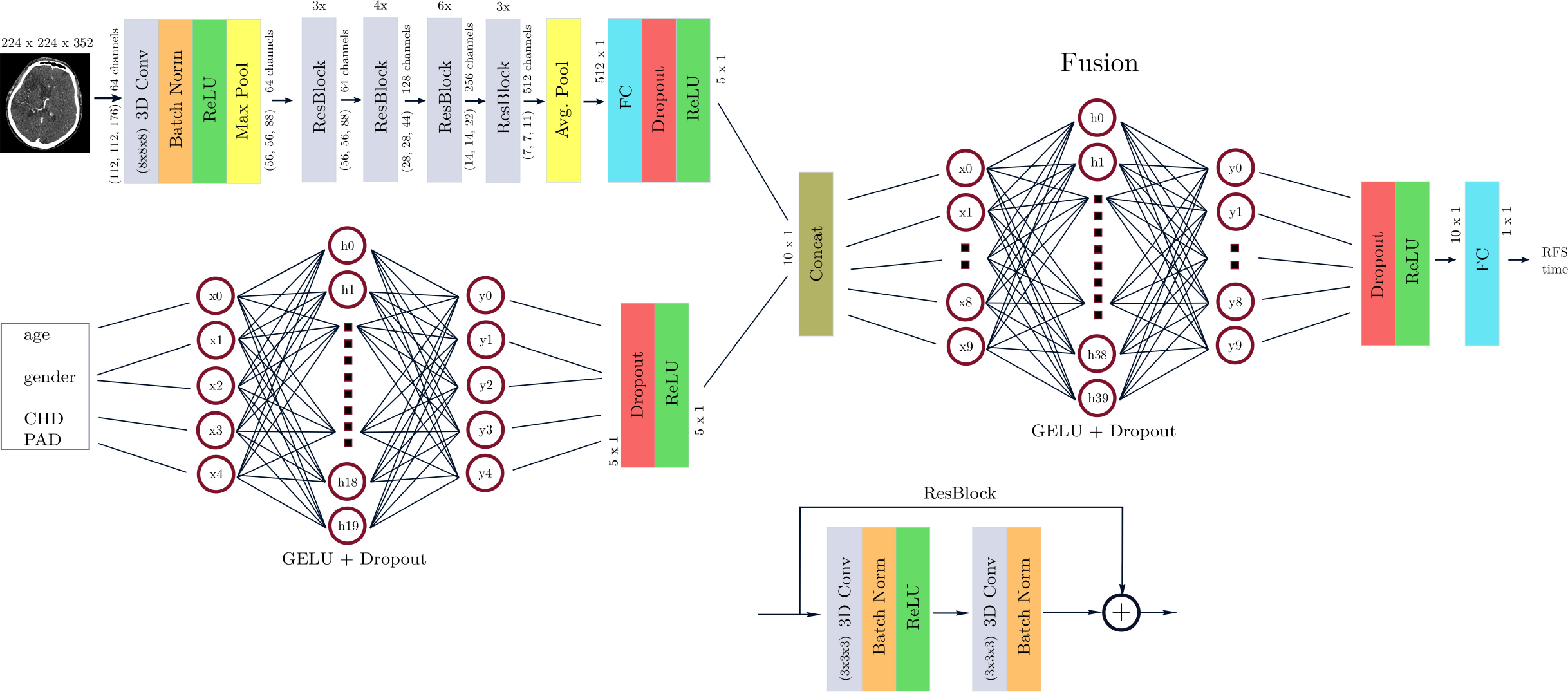}
	\caption{XSRD-net: Architecture. ResBlock: Residual Block, Conv: Convolution, Max Pool: Max Pooling, Avg. Pool: Average Pooling, ReLU: Rectified Linear Unit, GELU: Gaussian Error Linear Unit, FC: Fully Connected Layer, Concat: Concatenation.}
	\label{fig:XSRD}
\end{figure}

\paragraph*{Multimodal Neural Network}
Multimodal Fusion methods were tested in \cite{Gapp_Multimodal_Medical_Disease_Classification}, where experiments were carried out for Vision Transformers \cite{16x16WORDS} or ResNets \cite{ResNet} for vision, and LLaMA~II \cite{LLaMAII} for text data.
Based on the results in our prior work \cite{Gapp_Multimodal_Medical_Disease_Classification}, we decided to use a ResNet (ResNet34) for processing the 3D image data and a MLP (Multi-Layer Perceptron) for the tabular data.
The fusion model, viz. our designated XSRD-net (\cref{fig:XSRD}), combines the multimodal heterogeneous information with late concatenation, i.e., a late fusion method introduced in \cite{MultimodalTransformersSurvey}, and adds another MLP for further data processing.

\subsection{Experiments}
With Task~1 we train models to separate the relapse from non-relapse cases with binary classification as a first try. In the main task, Task~2, regression models to predict the RFS times are trained. A classification for the occurrence of relapses is done subsequently on the regression’s outcome. For both tasks we train single-modal (tabular-only, vision-only) and multimodal neural networks from scratch. The XSRD-net is the multimodal regression model for RFS time prediction. Training each task required approximately 10 hours on a NVIDIA TITAN RTX 24GB GPU, with implementations based on PyTorch \cite{Pytorch}. Performing high-resolution visual interpretability analysis took about 8 hours for one patient.

\section{Results}\label{sec:results}
The performance results for the classification, regression + subsequent classification tasks are presented below, followed by further interpretability.

\subsection{Task 1: Classification}
\begin{figure}[t!]
	\centering
	\subfigure[$\beta$  = 1.0]{\includegraphics[width=0.325\textwidth]{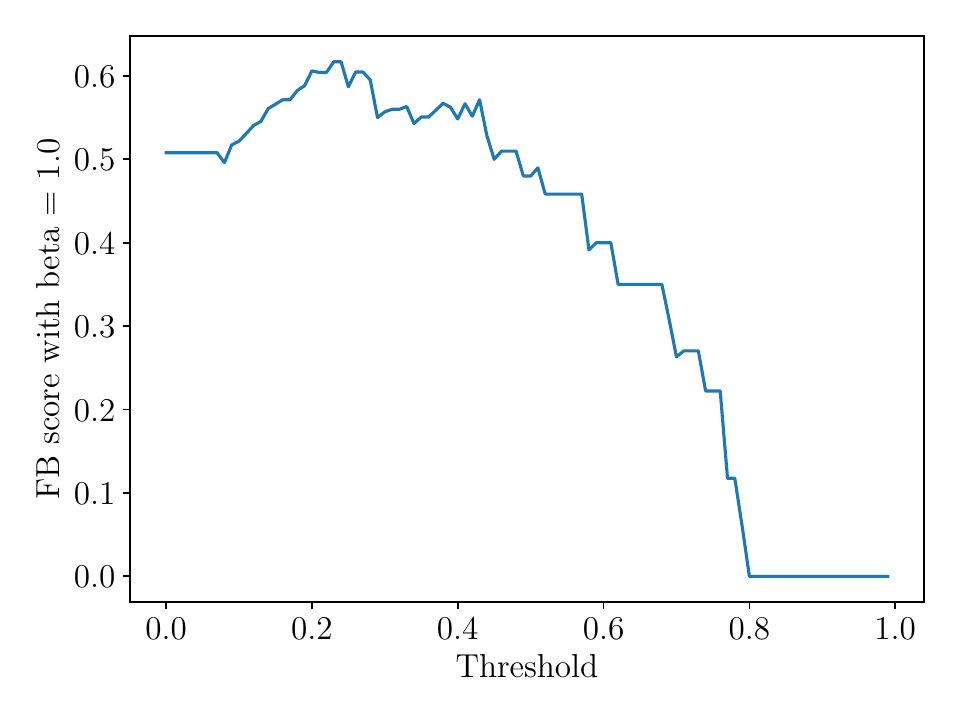}\centering}
	\subfigure[$\beta$ = 2.0]{\includegraphics[width=0.325\textwidth]{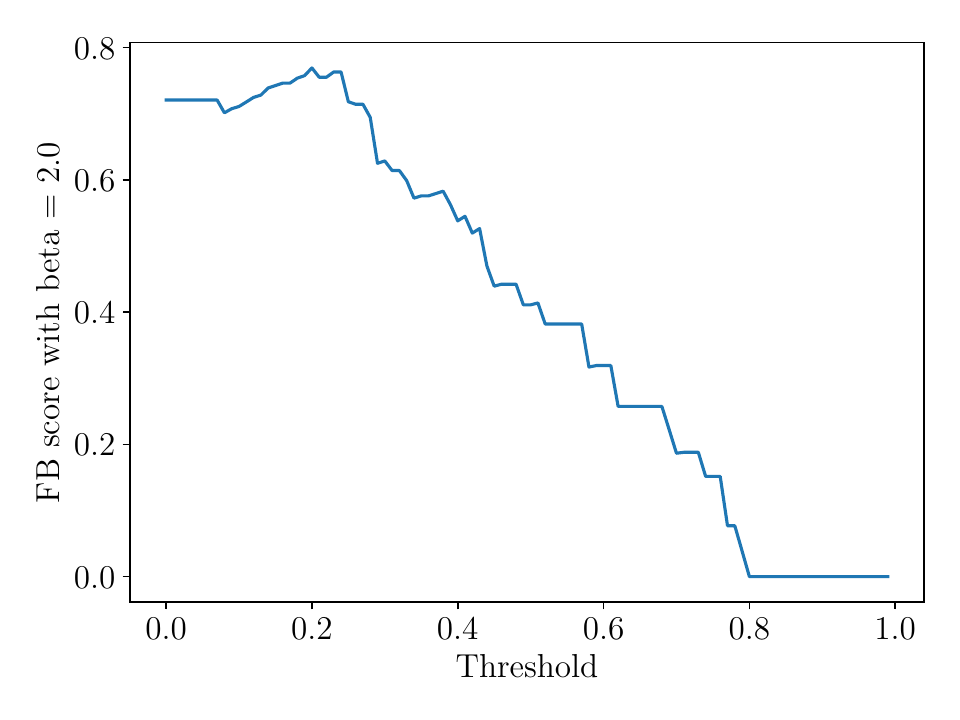}\centering}
	\subfigure[$\beta$ = 0.5]{\includegraphics[width=0.325\textwidth]{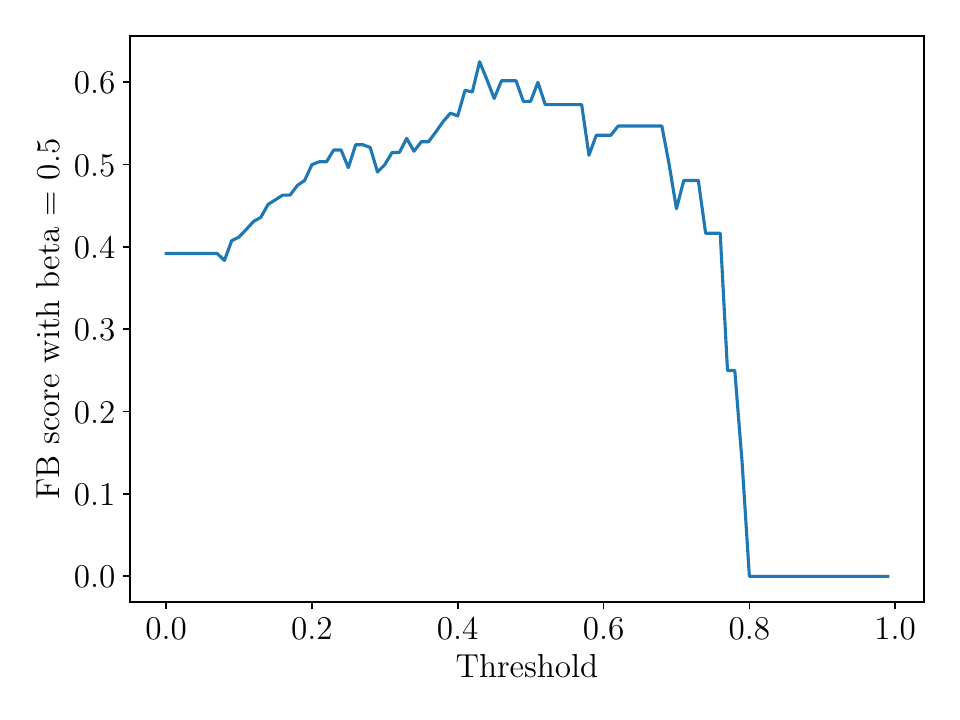}\centering}
	\caption{Tabular classification: F\textsubscript{$\beta$}-scores vs. thresholds with different values for $\beta$. Note that $\beta$~>~1 realizes a recall weighted, $\beta$~<~1 a precision weighted score. The best separating threshold is at the position, where the F\textsubscript{$\beta$} score reaches its maximum. The threshold selection is done on the train dataset and then kept fixed for the model.}
	\label{fig:tabularClassificaitonThresholds}
\end{figure}

A total overview of results for all models trained for classification is summarized in \cref{tab:classification_results_overview}. As the outcome of the classification is $\in$ [0.0, 1.0], a threshold $\theta$ must be set to separate the two classes, viz. relapse vs. relapse-free. Outcomes above $\theta$ are seen as relapses, outcomes below $\theta$ as relapse-free. 
The ideal threshold for binary classification is rarely exactly at 0.5. Due to data label imbalance it can be assumed that the ideal threshold for separating the data is $<$ 0.5 as the label 0 (no relapse) is more present in the data. As a criterion for the perfect threshold we take the F\textsubscript{1} score, viz. the harmonic mean between the recall (i.e. sensitivity) and precision. Note that one could also choose the F\textsubscript{$\beta$} score
\begin{equation*}
	\mathrm{F}_\beta = (1+ \beta^2)\frac{\text{precision}\cdot\text{recall}}{(\beta^2\cdot\text{precision})+\text{recall}} \ , \ \beta \in \mathbb{R}_0^+ \ ,
\end{equation*}
thus a recall ($\beta > 1$) or precision ($\beta < 1$) weighted score for threshold selection. \cref{fig:tabularClassificaitonThresholds} shows the F\textsubscript{$\beta$} scores vs. the thresholds for different values for $\beta$ using the tabular-only model.
In \cref{tab:tabularClassificationSensSpec} details to sensitivity and specificity are provided.

%
\setlength{\tabcolsep}{5pt}
\begin{table}[t!]
	\caption[Classification of relapses. Overview.]{Classification of relapses. Overview of performances of single-modal and multimodal classifications on the test dataset.}
	\centering
	\begin{tabular}{lccc}
		\toprule
		& tabular-only & vision-only & multimodal \\
		\cmidrule(lr){2-2}
		\cmidrule(lr){3-3}
		\cmidrule(lr){4-4}
		AUC &0.84 &0.67 &0.82 \\
		F\textsubscript{1} &0.69 &0.53 &0.69 \\
		\bottomrule
		\label{tab:classification_results_overview}
	\end{tabular}
\end{table}

\begin{table}[t!]
	\caption[Classification metrics with different thresholds]{Classification results, viz. sensitivity (Sens.) and specificity (Spec.), for the test dataset with different thresholds $\theta \in \mathbb{R} \cap [0,1]$ using the tabular-only model. The first threshold maximizes F\textsubscript{1} score, the second one maximizes F\textsubscript{2} score, the last one maximizes F\textsubscript{0.5} score for the train dataset.}
	\centering
	\begin{tabular}{lcccccc}
		\toprule
		$\theta$& \multicolumn{2}{c}{0.23} & \multicolumn{2}{c}{0.2}  & \multicolumn{2}{c}{0.43}\\
		\cmidrule(lr){2-3} 
		\cmidrule(lr){4-5} 
		\cmidrule(lr){6-7}
		& Sens.& Spec.& Sens.& Spec.& Sens.& Spec.\\
		\cmidrule(lr){2-2} 
		\cmidrule(lr){3-3}
		\cmidrule(lr){4-4} 
		\cmidrule(lr){5-5} 
		\cmidrule(lr){6-6}
		\cmidrule(lr){7-7} 
		test &1.00 &0.50 &1.00&0.50&0.78&0.81\\
		\bottomrule
		\label{tab:tabularClassificationSensSpec}
	\end{tabular}
\end{table}

\subsection{Task 2: Regression}
\begin{figure}[t!]
	\centering
	\includegraphics[width=0.35\textwidth]{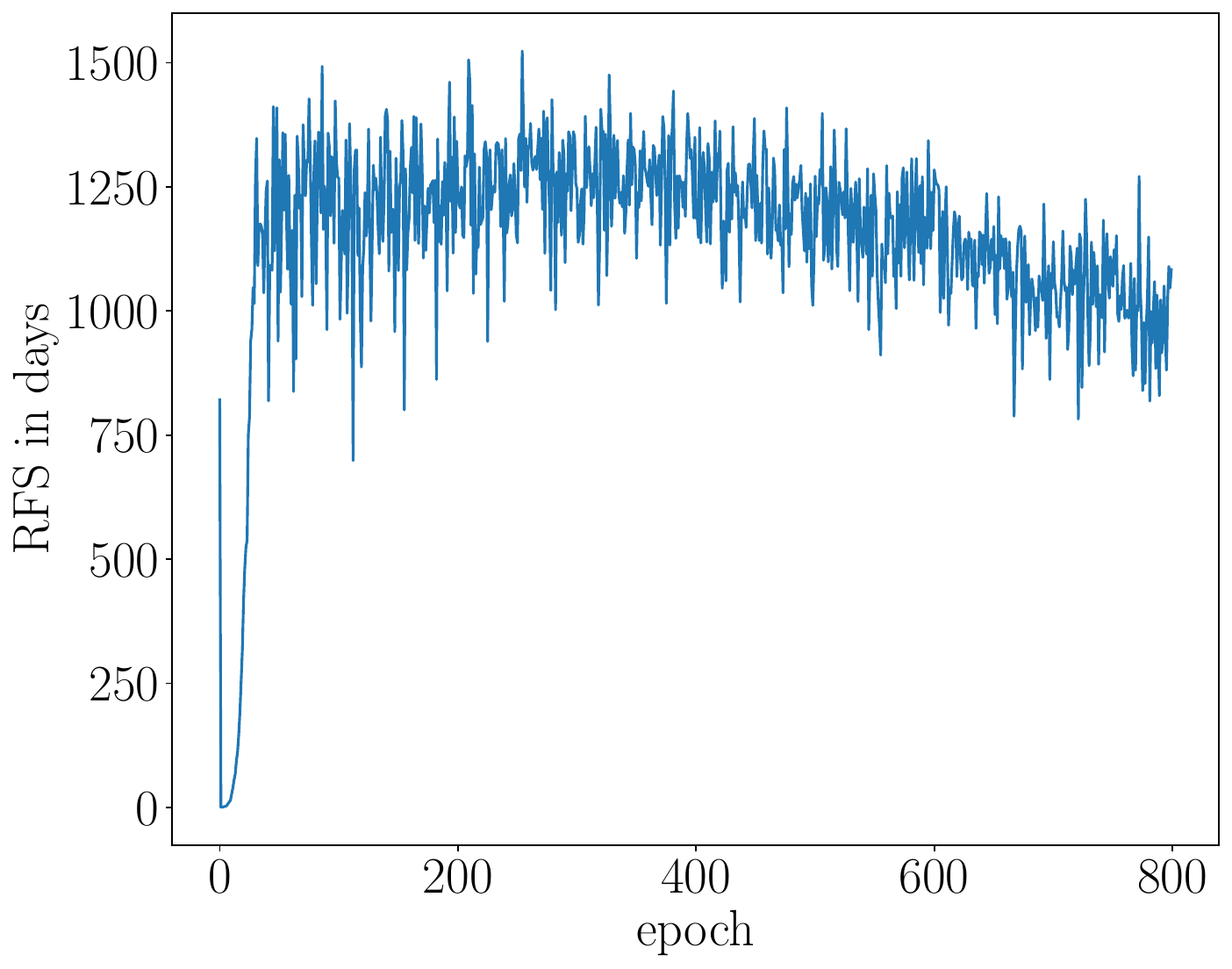}
	\caption[Progression of RFS estimation value]{Progression of RFS estimation value for multimodal regression. The plot shows the mean value progression of the predicted RFS times of those relapse patients, which have a known ground truth RFS. This mean value is used during training to initialize the ground truths of the relapses with unknown RFS due to missing data. In later epochs the RFS prediction approaches the mean of the ground truth RFS from relapse patients.}
	\label{fig:rfs_estimation_progression}
\end{figure}

The regression training for predicting the RFS times is done for all patients independently of the labels. The performance analysis is partly done with respect to the labels (AUC, F\textsubscript{1}). For the evaluation of patients with missing ground truth rfs times in the testing dataset, $1/2\cdot \kappa$\textsubscript{low} was used. The results are depicted in \cref{tab:regression_results_overview}. 
%
\cref{fig:rfs_estimation_progression} shows the RFS time estimation for incomplete data as described in \cref{para:handlingIncompleteData}.

\begin{table}[t!!]
	\caption[Regression of RFS. Overview.]{Regression of RFS. Overview of performances of single-modal and multimodal (XSRD-net) regressions on the test dataset. Additionally the classification results (AUC, F\textsubscript{1}) are provided for the subsequent classification task.}
	\centering
	\begin{tabular}{lccc}
		\toprule
		& tabular-only & vision-only & XSRD-net \\
		\cmidrule(lr){2-2}
		\cmidrule(lr){3-3}
		\cmidrule(lr){4-4}
		c-index &0.45 &0.59 &0.56\\
		c-index relapses &0.38 &0.64 &0.68\\
		AUC &0.34 &0.72 &0.71\\
		F\textsubscript{1} &0.53 &0.60 &0.63 \\
		\bottomrule
		\label{tab:regression_results_overview}
	\end{tabular}
\end{table}

\subsubsection*{Fusion: Regression + Classification}
\begin{figure}[t!]
	\centering
	\subfigure[$\beta$  = 1.0]{\includegraphics[width=0.32\textwidth]{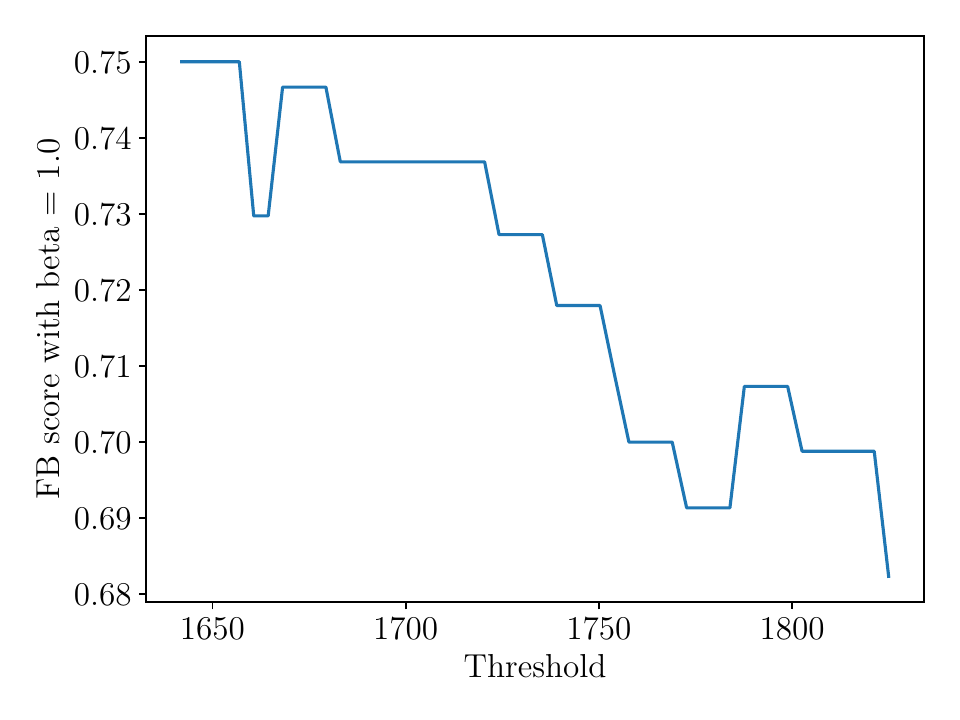}}
	\subfigure[$\beta$  = 2.0]{\includegraphics[width=0.32\textwidth]{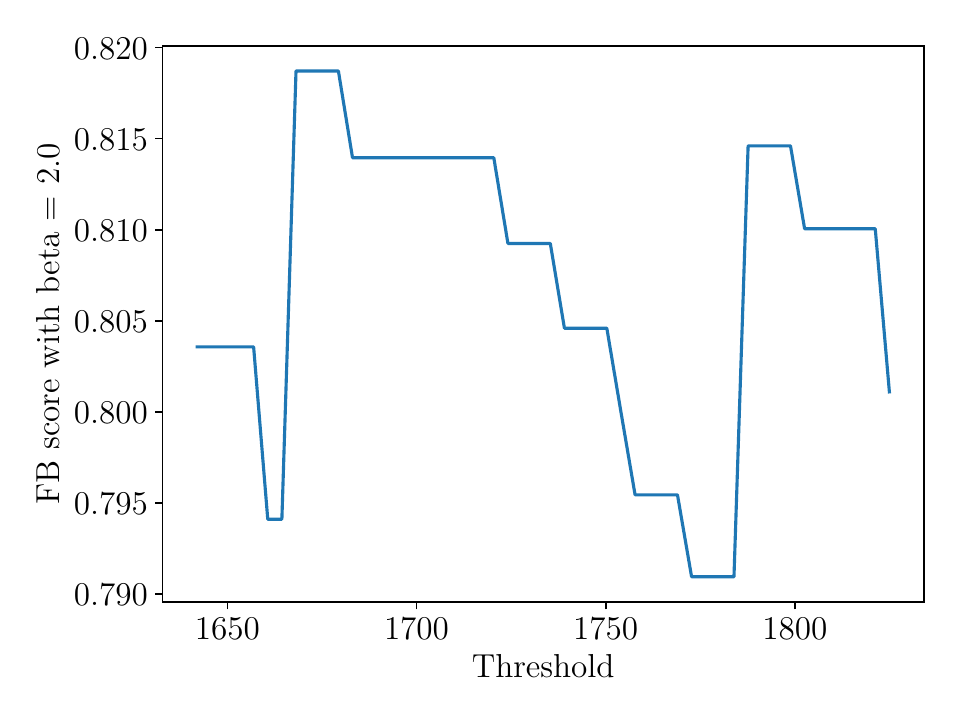}}
	\subfigure[$\beta$  = 0.5]{\includegraphics[width=0.32\textwidth]{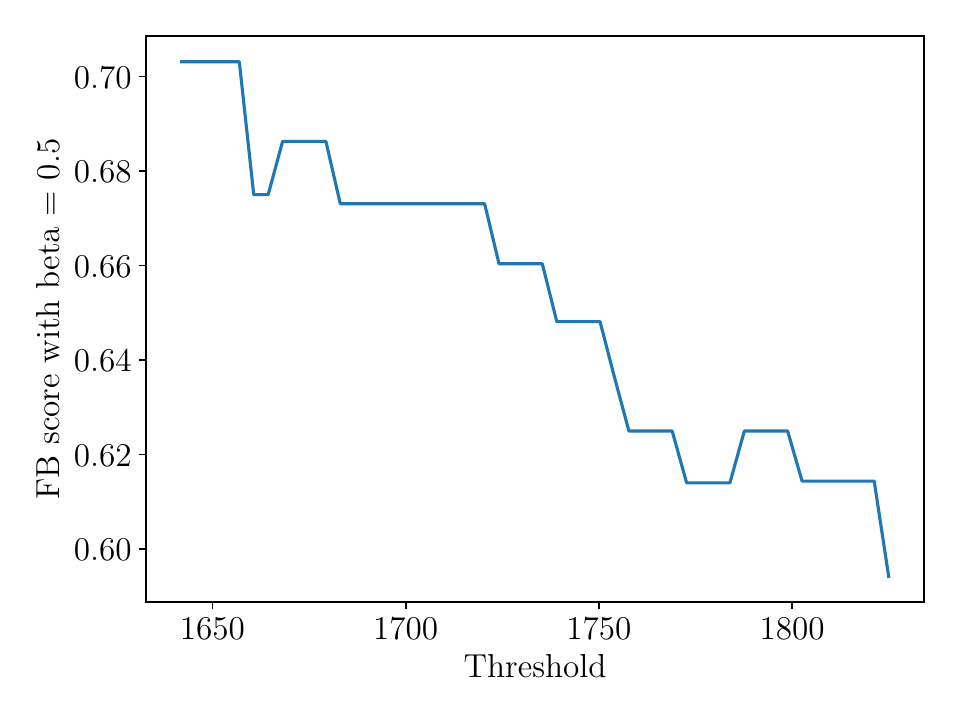}}
	\caption{Multimodal regression: F\textsubscript{$\beta$}-scores vs. thresholds for classification after regression with different values for $\beta$. Note that $\beta$~>~1 realizes a recall weighted, $\beta$~<~1 a precision weighted score. The position, where the F\textsubscript{$\beta$} score reaches its maximum, is the best separating threshold. The selection of this threshold is done on the train dataset and then kept fixed for the model.}
	\label{fig:RegressionClassification_f1_vs_threshold}
\end{figure}

Since we have separated the data with a maximum RFS for relapses ($\kappa$\textsubscript{low}), and a minimum RFS for non-relapse cases ($\kappa$\textsubscript{high} > $\kappa$\textsubscript{low}), we can do a classification after the regression. 
%
For this, we take a threshold $\kappa$ in days between these two values ($\kappa$\textsubscript{low} $\le$ $\kappa$ $\le$ $\kappa$\textsubscript{hiqh}). Predicted RFS times $\le \kappa$ are classified as relapses, others as non-relapses. $\kappa$ near $\kappa$\textsubscript{low} increases specificity to the cost of FNs, $\kappa$ near $\kappa$\textsubscript{high} increases sensitivity to the cost of FPs. Analogously to the binary classification (Task~1), we choose $\kappa$ as the best separating value, where the F\textsubscript{1} score reaches the maximum. 
Details see \cref{tab:classificationAfterRegressionThreshold} and \cref{fig:RegressionClassification_f1_vs_threshold}.
The resulting performances are summarized in \cref{tab:regression_results_overview} for all models.

\begin{table}[t!]
	\caption[Classification metrics with different thresholds]{Classification results, viz. sensitivity (Sens.) and specificity (Spec.), for the test dataset with different thresholds $\kappa$ in days using the multimodal model -- XSRD-net. The first threshold maximizes F\textsubscript{1} score, the second one maximizes F\textsubscript{2} score, the third one is for maximization of F\textsubscript{4} score and the last one maximizes F\textsubscript{0.5} score for the train dataset.}
	\centering
	\begin{tabular}{lcccccccc}
		\toprule
		$\kappa$& \multicolumn{2}{c}{1642.00} & \multicolumn{2}{c}{1668.14}  & \multicolumn{2}{c}{1787.65} &\multicolumn{2}{c}{1642.00}\\
		\cmidrule(lr){2-3} 
		\cmidrule(lr){4-5} 
		\cmidrule(lr){6-7}
		\cmidrule(lr){8-9}
		& Sens.& Spec.& Sens.& Spec.& Sens.& Spec.& Sens.& Spec.\\
		\cmidrule(lr){2-2} 
		\cmidrule(lr){3-3}
		\cmidrule(lr){4-4} 
		\cmidrule(lr){5-5} 
		\cmidrule(lr){6-6}
		\cmidrule(lr){7-7}
		\cmidrule(lr){8-8}
		\cmidrule(lr){9-9}
		test &0.56&0.88&0.56&0.88&0.56&0.75&0.56&0.88\\
		\bottomrule
	\end{tabular}
	\label{tab:classificationAfterRegressionThreshold}
\end{table}

\subsection{Intepretability}

\paragraph*{Modality Contribution} The modality contribution metric $m_{i} \in \mathbb{R} \cap [0,1]$, with $\sum_{i=1}^{N}m_i = 1.0$, is a measure for how much the model uses modality $i \in \{1, \dots, N\}$ ($N$: number of modalities) for solving a task \cite{Gapp_MCI}.
For the multimodal tasks we used vision data (modality 1) and tabular data (modality 2). The tabular data entries, viz. age, gender, CHD and PAD, are analyzed deeper with the same method. This gives us an insight into the importance of the four attributes itself.
Results are presented in \cref{tab:mci_results_classification} for the classification and the regression tasks.

\begin{table}[t!]
	\caption[Modality contributions for classification of relapses.]{Modality contributions. Top: binary classification. Bottom: RFS prediction. Values: absolute importance. The sum of these values is 1.0 for each row. Values in parentheses: relative importance per modality. The sum of these values is 1.0 for each modality. The vision modality consists of one attribute, viz. the whole image. For the tabular modality the relative importance of each attribute is provided additionally.}
	\centering
	\begin{tabular}{lccccc}
		\toprule
		& vision & \multicolumn{4}{c}{tabular}\\
		\cmidrule(lr){2-2}
		\cmidrule(lr){3-6}		
		& 3D CTA & age & gender & CHD & PAD \\
		\cmidrule(lr){2-2}
		\cmidrule(lr){3-3}
		\cmidrule(lr){4-4}
		\cmidrule(lr){5-5}
		\cmidrule(lr){6-6}
		tabular-only &- & 0.46 (0.46) &0.35 (0.35) & 0.06 (0.06) &0.13 (0.13)\\
		vision-only & 1.00 (1.00) & - &- &- &-\\
		multimodal &0.01 (1.00) &0.47 (0.48) &0.35 (0.35) &0.07 (0.07) &0.10 (0.10) \\
		\bottomrule
		\label{tab:mci_results_classification}
	\end{tabular}
	\begin{tabular}{lccccc}
		\toprule
		& vision & \multicolumn{4}{c}{tabular}\\
		\cmidrule(lr){2-2}
		\cmidrule(lr){3-6}	
		& 3D CTA & age & gender & CHD & PAD \\
		\cmidrule(lr){2-2}
		\cmidrule(lr){3-3}
		\cmidrule(lr){4-4}
		\cmidrule(lr){5-5}
		\cmidrule(lr){6-6}
		tabular-only &- &0.05 (0.05) &0.80 (0.80) &0.10 (0.10) &0.05 (0.05)\\
		vision-only & 1.00 (1.00) & - &- &- &-\\
		XSRD-net & 0.68 (1.00) & 0.03 (0.09) &0.20 (0.63)  &0.03 (0.09) &0.06 (0.19) \\
		\bottomrule
		\label{tab:mci_results_regression}
	\end{tabular}
\end{table}

\begin{figure}[t!]
	\centering
	\begin{minipage}[b]{0.48\textwidth}
		\centering
		\includegraphics[width=0.22\textwidth]{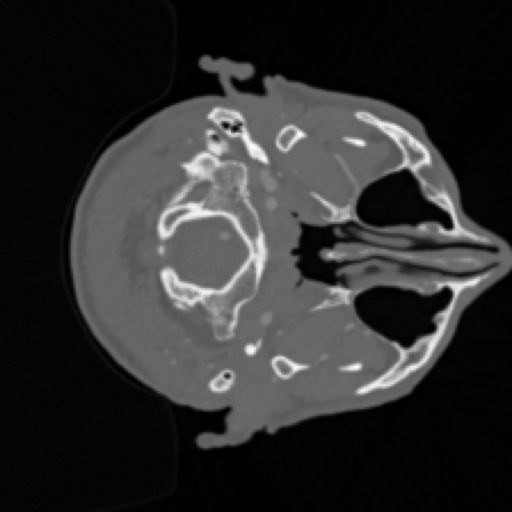}
		\includegraphics[width=0.3452\textwidth]{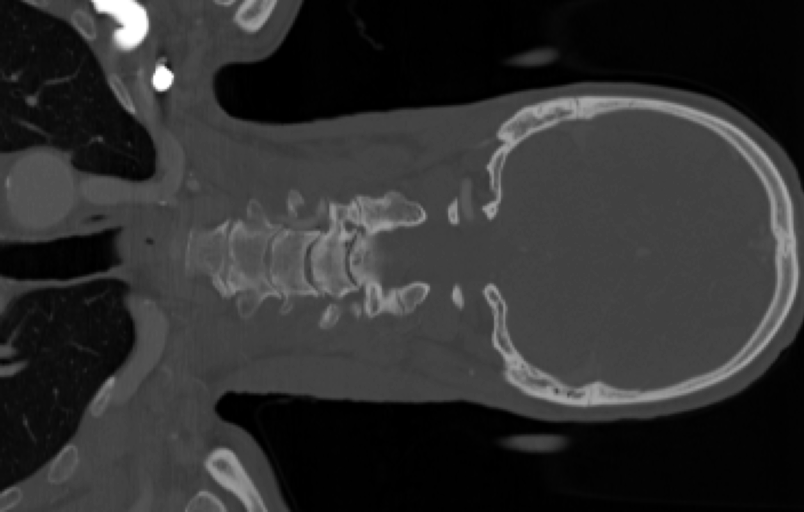}
		\includegraphics[width=0.3452\textwidth]{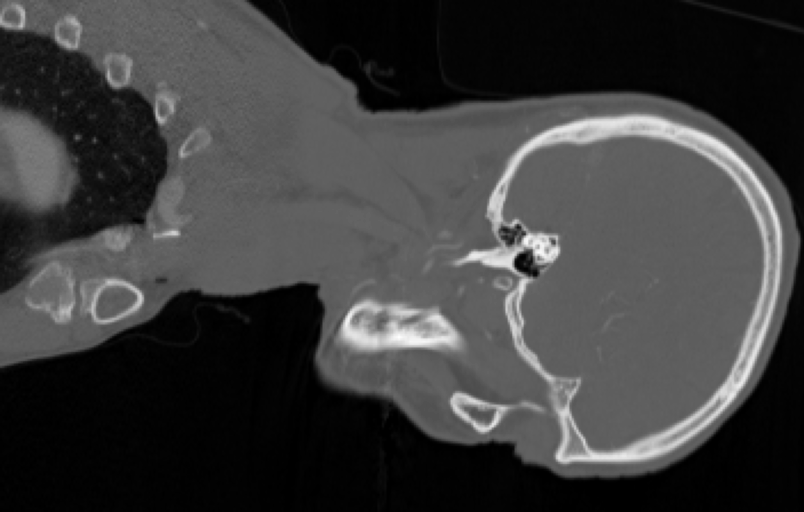}
		\includegraphics[width=0.22\textwidth]{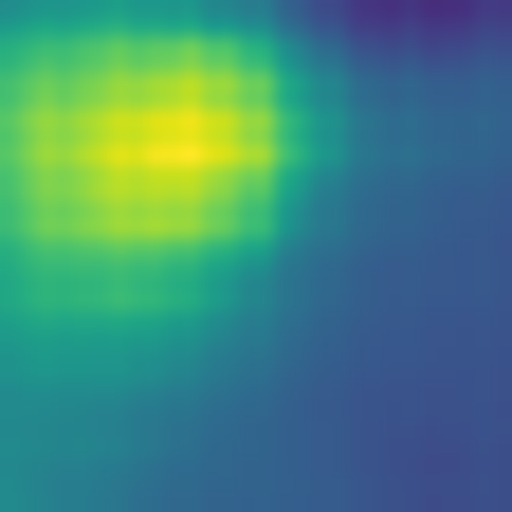}
		\includegraphics[width=0.3452\textwidth]{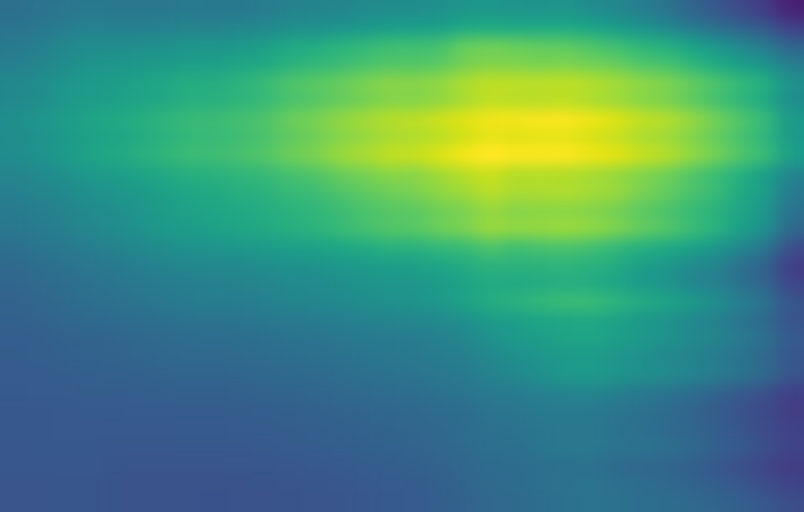}
		\includegraphics[width=0.3452\textwidth]{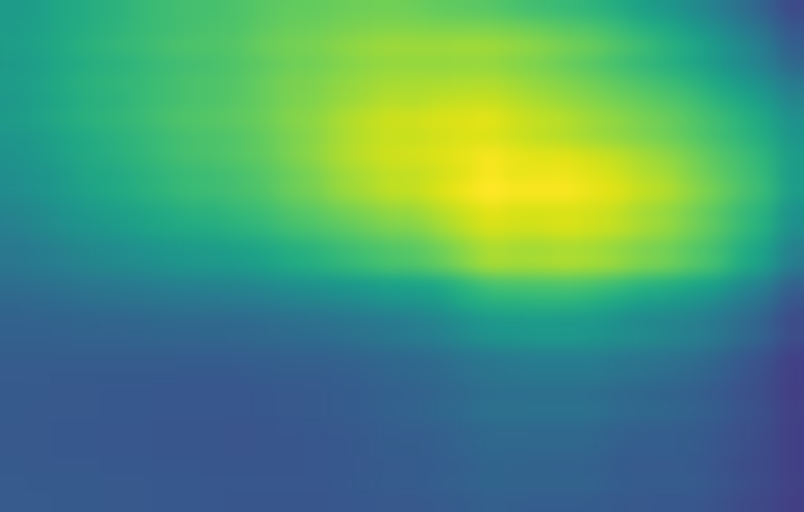}
	\end{minipage}
	\begin{minipage}[b]{0.48\textwidth}
		\centering
		\includegraphics[width=0.22\textwidth]{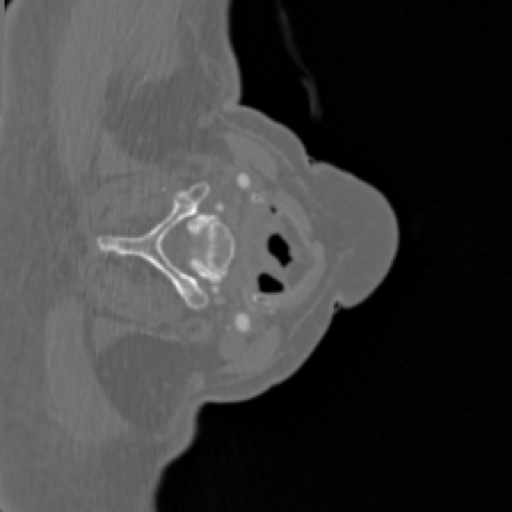}
		\includegraphics[width=0.3452\textwidth]{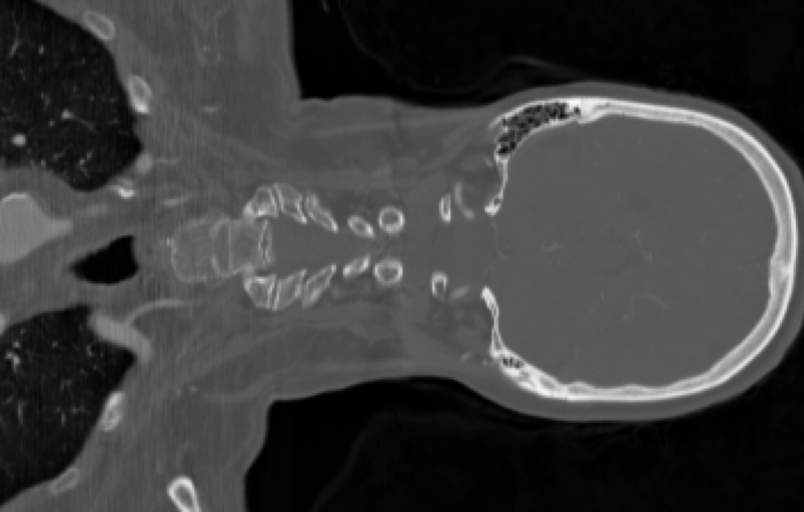}
		\includegraphics[width=0.3452\textwidth]{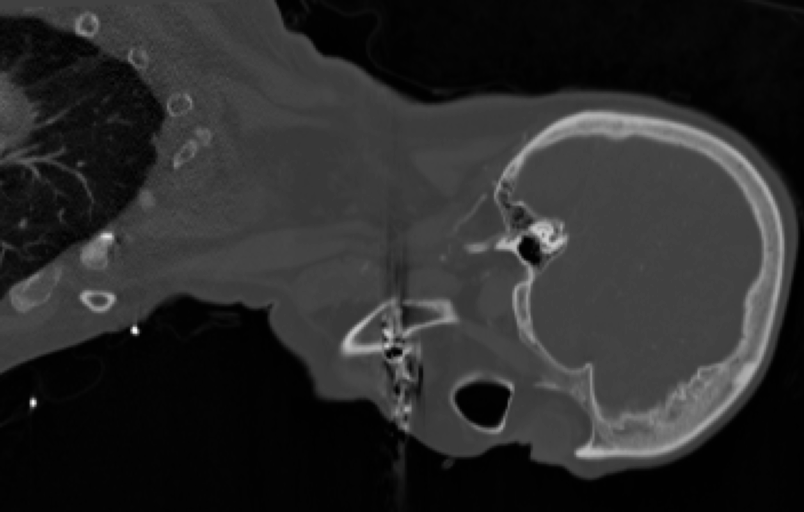}
		\includegraphics[width=0.22\textwidth]{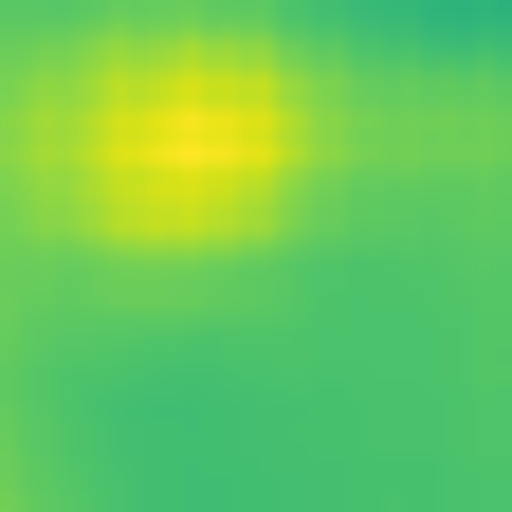}
		\includegraphics[width=0.3452\textwidth]{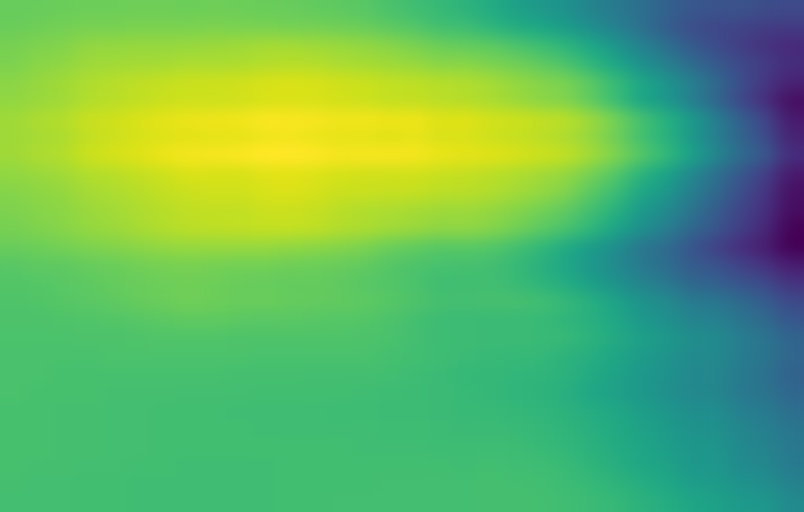}
		\includegraphics[width=0.3452\textwidth]{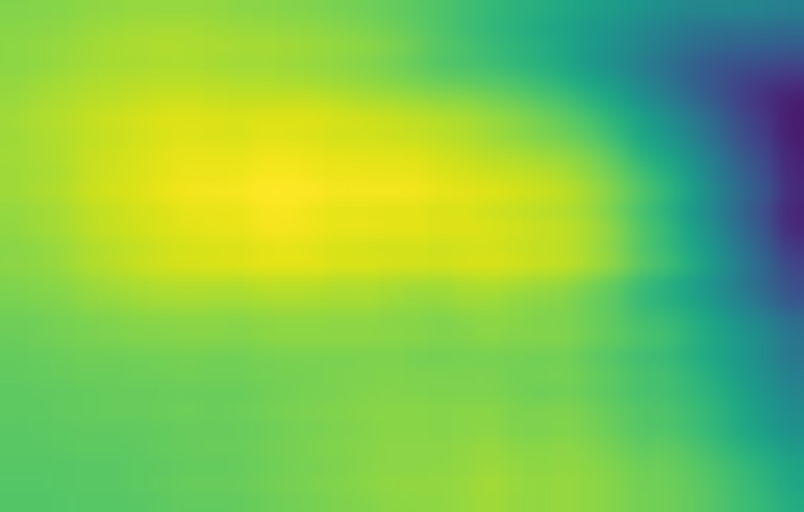}
	\end{minipage}
	\caption{Occlusion sensitivity on 3D CTAs with the XSRD-net. Left: TN example (RFS = 2372 days). Right: TP example (RFS = 107 days). 
		Top row, from left to right: transversal slice (viewed from top), coronal slice (viewed from back), sagittal slice (viewed from left). Bottom row: saliency maps for the slices. Yellow regions mark importance.}
	\label{fig:vision_interpretability}
\end{figure}

\paragraph{Vision Interpretability}
Results of a detailed occlusion sensitivity based interpretability analysis are shown in \cref{fig:vision_interpretability}. The visual region of interest in the CTAs for the RFS time prediction is the supra-aortic area up to the upper neck, the region where the arteria carotis communis is located.

\section{Discussion}\label{sec:discussion}

For Task~1, the classification task, the best results (see \cref{tab:classification_results_overview}) are reached with tabular data: AUC 0.84 for the test dataset. The vision-only model could not reliably distinguish between relapse and non-relapse patients (AUC: 0.67 (test)). The multimodal model performs similar to the single-modal tabular-only model. Interpretability results show that an unimodal collapse occurred for the multimodal classification network.
Modality contribution results for tabular classification (\cref{tab:mci_results_classification}) further show that age (0.46) and gender (0.35) are the most important attributes to detect relapses. However, heart diseases have a remarkable modality contribution value of 0.19 (0.06 (CHD) + 0.13 (PAD)). 

Task~2, the regression task, is best performed by the vision-only (c-index: 0.59 (test)) and the multimodal XSRD-net (c-index: 0.56 (test)) -- see \cref{tab:regression_results_overview}.
With respect to the relapse cases, the c-index is 0.64 (vision-only) and 0.68 (multimodal) for the test dataset. That means for relapses, both models can predict the RFS time more precise than for all cases. 
%
The subsequent classification applied on the regression outcome reaches an AUC of 0.71 and a F\textsubscript{1}-score of 0.63 for the test dataset with the XSRD-net (\cref{tab:regression_results_overview}).
Tabular data was used 32\%, vision 68\% (\cref{tab:mci_results_regression}). Despite the tabular-only model's poor performance, the multimodal model still benefits from late fusion of vision with tabular data.

The overall contribution of the tabular data in the multimodal regression task for age is 0.03, gender:~0.20, CHD:~0.03, PAD:~0.06.  
The comparison of the relative attributes' importance from multimodal regression with that from tabular classification (see \cref{tab:mci_results_regression}) results in age:~$-$0.37, gender: +0.28, CHD: +0.03, PAD: +0.06.
The gender is the most important attribute within the tabular data for regression. However, pre-existing diseases (CHD and PAD) are still essential markers.
Occlusion sensitivity based analysis of the image data in \cref{fig:vision_interpretability} show that the supra-aortal region up to the upper neck with the carotid arteries is of significant interest for RFS prediction. As dissections of the carotid arteries can lead to stroke events \cite{UKDCarotis}, it is obvious that plaque, vascular courses and shapes have a contribution to recurrences of stroke (see also \cite{Van_Dam-Nolen2022-qo,Singh2013}).

\section{Summary and Outlook}\label{sec:conclusion}

In order to support early relapse detection, we analyzed 3D CTA and clinical data (age, gender, CHD, PAD) from 119 ICE patients, including 41 relapses.
%
We trained single-modal and multimodal neural networks for binary classification of relapses (Task~1) and the prediction of RFS times (Task~2). Additionally we applied a subsequent classification on the regression task.

For Task~1 the tabular data was sufficient to separate the non-relapse from the relapse patients with an AUC of 0.84 for the test dataset. That is why an unimodal collapse occurred with the multimodal model.
In contrast, for Task~2 the XSRD-net processed the multimodal data more balanced (0.68 (vision) vs. 0.32 (tabular)). For the test datasets the model reached a c-index/c-index\textsubscript{(relapses)} of 0.56/0.68.
Regarding the subsequent classification applied on Task~2, the multimodal model reaches an AUC of 0.71 on the test dataset. 

Deeper modality contribution analysis \cite{Gapp_MCI} shows that for all tasks, the heart conditions CHD and PAD significantly influence outcomes. 
However, 3D CTA images are especially relevant for RFS prediction. Occlusion sensitivity analysis highlights the supra-aortal region up to the upper neck (arteria carotis communis) as most relevant, particularly in correctly predicted cases (TP and TN). This aligns with \cite{Singh2013}, which found a strong link between stroke recurrence and echolucent carotid plaques. 
With continued data collection and model updates, we aim to further explore the carotid-stroke relationship to identify new biomarkers.

{\small
\subsubsection{\small\ackname}
 This study is partly supported by VASCage -- Centre on Clinical Stroke Research.

\subsubsection{\small Disclosure of Interests.}
The authors have no competing interests to declare that are
relevant to the content of this article.

%
%
\bibliographystyle{spmpsci}
\bibliography{references}

\begin{thebibliography}{10}
\providecommand{\url}[1]{{#1}}
\providecommand{\urlprefix}{URL }
\expandafter\ifx\csname urlstyle\endcsname\relax
  \providecommand{\doi}[1]{DOI~\discretionary{}{}{}#1}\else
  \providecommand{\doi}{DOI~\discretionary{}{}{}\begingroup
  \urlstyle{rm}\Url}\fi

\bibitem{Callaly2016-dq}
Callaly, E., Ni~Chroinin, D., Hannon, N., Marnane, M., Akijian, L., Sheehan,
  O., Merwick, A., Hayden, D., Horgan, G., Duggan, J., Murphy, S., O'Rourke,
  K., Dolan, E., Williams, D., Kyne, L., Kelly, P.J.: Rates, predictors, and
  outcomes of early and late recurrence after stroke: The {N}orth {D}ublin
  population stroke study.
\newblock Stroke \textbf{47}(1), 244--246 (2016).
\newblock \doi{10.1161/STROKEAHA.115.011248}

\bibitem{Caplan_2016}
Caplan, L.R. (ed.): Caplan’s Stroke: A Clinical Approach, 5th edn.
\newblock Cambridge University Press (2016).
\newblock \doi{10.1017/CBO9781316095805}

\bibitem{Chung2023-nt}
Chung, J.Y., Lee, B.N., Kim, Y.S., Shin, B.S., Kang, H.G.: Sex differences and
  risk factors in recurrent ischemic stroke.
\newblock Front Neurol \textbf{14}, 1028,431 (2023).
\newblock \doi{10.3389/fneur.2023.1028431}

\bibitem{Van_Dam-Nolen2022-qo}
van Dam-Nolen, D.H.K., Truijman, M.T.B., van~der Kolk, A.G., Liem, M.I.,
  Schreuder, F.H.B.M., Boersma, E., Daemen, M.J.A.P., Mess, W.H., van
  Oostenbrugge, R.J., van~der Steen, A.F.W., Bos, D., Koudstaal, P.J.,
  Nederkoorn, P.J., Hendrikse, J., van~der Lugt, A., Kooi, M.E., {PARISK Study
  Group}: Carotid plaque characteristics predict recurrent ischemic stroke and
  {TIA}: The {PARISK} (plaque at {RISK}) study.
\newblock JACC Cardiovasc. Imaging \textbf{15}(10), 1715--1726 (2022).
\newblock \doi{10.1016/j.jcmg.2022.04.003}

\bibitem{Ding2024}
Ding, X., Meng, Y., Xiang, L., Boden-Albala, B.: Stroke recurrence prediction
  using machine learning and segmented neural network risk factor aggregation.
\newblock Discover Public Health \textbf{21}(1), 119 (2024).
\newblock \doi{10.1186/s12982-024-00199-6}

\bibitem{Donkor2018-zg}
Donkor, E.S.: Stroke in the 21(st) century: A snapshot of the burden,
  epidemiology, and quality of life.
\newblock Stroke Res Treat \textbf{2018}, 3238,165 (2018).
\newblock \doi{10.1155/2018/3238165}

\bibitem{16x16WORDS}
Dosovitskiy, A., Beyer, L., Kolesnikov, A., Weissenborn, D., Zhai, X.,
  Unterthiner, T., Dehghani, M., Minderer, M., Heigold, G., Gelly, S.,
  Uszkoreit, J., Houlsby, N.: An image is worth 16x16 words: Transformers for
  image recognition at scale.
\newblock In: International Conference on Learning Representations (2021).
\newblock \doi{10.48550/arxiv.2010.11929}

\bibitem{MultimodalStrokeModel2024}
Fan, D., Miao, R., Huang, H., Wang, X., Li, S., Huang, Q., Yang, S., Deng, R.:
  Multimodal ischemic stroke recurrence prediction model based on the capsule
  neural network and support vector machine.
\newblock Medicine \textbf{103}(35) (2024).
\newblock \doi{10.1097/MD.0000000000039217}

\bibitem{Gapp_MCI}
Gapp, C., Tappeiner, E., Welk, M., Fritscher, K., Gizewski, E.R., Schubert, R.:
  What are you looking at? {M}odality contribution in multimodal medical deep
  learning methods  (2025).
\newblock \doi{10.48550/arXiv.2503.01904}.
\newblock {Conference for Computer Assisted Radiology and Surgery (CARS) -- (in
  press)}

\bibitem{Gapp_Multimodal_Medical_Disease_Classification}
Gapp, C., Tappeiner, E., Welk, M., Schubert, R.: Multimodal medical disease
  classification with {LLaMA~II}  (2024).
\newblock \doi{10.48550/arXiv.2412.01306}.
\newblock {The First Austrian Symposium on AI, Robotics, and Vision (AIROV24)}

\bibitem{ResNet}
He, K., Zhang, X., Ren, S., Sun, J.: Deep residual learning for image
  recognition.
\newblock In: 2016 IEEE Conference on Computer Vision and Pattern Recognition
  (CVPR), pp. 770--778 (2016).
\newblock \doi{10.1109/CVPR.2016.90}

\bibitem{StrokeOutcomeMultimodal23}
Liu, Y., Yu, Y., Ouyang, J., Jiang, B., Yang, G., Ostmeier, S., Wintermark, M.,
  Michel, P., Liebeskind, D.S., Lansberg, M.G., Albers, G.W., Zaharchuk, G.:
  Functional outcome prediction in acute ischemic stroke using a fused imaging
  and clinical deep learning model.
\newblock Stroke \textbf{54}(9), 2316--2327 (2023).
\newblock \doi{10.1161/STROKEAHA.123.044072}

\bibitem{SimpleElastics}
Marstal, K., Berendsen, F., Staring, M., Klein, S.: Simpleelastix: A
  user-friendly, multi-lingual library for medical image registration.
\newblock In: 2016 IEEE Conference on Computer Vision and Pattern Recognition
  Workshops (CVPRW), pp. 574--582 (2016).
\newblock \doi{10.1109/CVPRW.2016.78}

\bibitem{Murphy2020-rf}
Murphy, S.J., Werring, D.J.: Stroke: causes and clinical features.
\newblock Medicine (Abingdon) \textbf{48}(9), 561--566 (2020).
\newblock \doi{10.1016/j.mpmed.2020.06.002}

\bibitem{Pytorch}
Paszke, A., Gross, S., Massa, F., Lerer, A., Bradbury, J., Chanan, G., Killeen,
  T., Lin, Z., Gimelshein, N., Antiga, L., Desmaison, A., Kopf, A., Yang, E.,
  DeVito, Z., Raison, M., Tejani, A., Chilamkurthy, S., Steiner, B., Fang, L.,
  Bai, J., Chintala, S.: Pytorch: An imperative style, high-performance deep
  learning library.
\newblock In: Advances in Neural Information Processing Systems, vol.~32
  (2019).
\newblock \doi{10.48550/arXiv.1912.01703}

\bibitem{Peng2022-zj}
Peng, Y., Ngo, L., Hay, K., Alghamry, A., Colebourne, K., Ranasinghe, I.:
  {L}ong-term survival, stroke recurrence, and life expectancy after an acute
  stroke in {A}ustralia and {New Zealand} from 2008--2017: A population-wide
  cohort study.
\newblock Stroke \textbf{53}(8), 2538--2548 (2022).
\newblock \doi{10.1161/STROKEAHA.121.038155}

\bibitem{Singh2013}
Singh, A.S., Atam, V., Jain, N., Yathish, B.E., Patil, M.R., Das, L.:
  Association of carotid plaque echogenicity with recurrence of ischemic
  stroke.
\newblock N. Am. J. Med. Sci. \textbf{5}(6), 371--376 (2013).
\newblock \doi{10.4103/1947-2714.114170}

\bibitem{LLaMAII}
Touvron, H., Martin, L., Edunov, S., Scialom, T.: Llama 2: Open foundation and
  fine-tuned chat models  (2023).
\newblock \doi{10.48550/arXiv.2307.09288}

\bibitem{UKDCarotis}
{University Hospital Carl Gustav Carus Dresden}: Treatment spectrum: Carotid
  stenosis.
\newblock
  \urlprefix\url{https://www.uniklinikum-dresden.de/de/das-klinikum/kliniken-polikliniken-institute/vtg/gefaesschirurgie/Behandlungsspektrum/carotisstenose}.
\newblock Accessed: 2025-05-22

\bibitem{Uzuner2023-gz}
Uzuner, N., Uzuner, G.T.: Risk factors for multiple recurrent ischemic strokes.
\newblock Brain Circ. \textbf{9}(1), 21--24 (2023).
\newblock \doi{10.4103/bc.bc_73_22}

\bibitem{Wang2023-ti}
Wang, Y., Fan, H., Duan, W., Ren, Z., Liu, X., Liu, T., Li, Y., Zhang, K., Fan,
  H., Ren, J., Li, J., Li, X., Wu, X., Niu, X.: Elevated stress hyperglycemia
  and the presence of intracranial artery stenosis increase the risk of
  recurrent stroke.
\newblock Front Endocrinol (Lausanne) \textbf{13}, 954,916 (2023).
\newblock \doi{10.3389/fendo.2022.954916}

\bibitem{MultimodalTransformersSurvey}
Xu, P., Zhu, X., Clifton, D.A.: Multimodal learning with transformers: A
  survey.
\newblock {IEEE} Transactions on Pattern Analysis \& Machine Intelligence
  \textbf{45}(10), 12,113--12,132 (2023).
\newblock \doi{10.1109/TPAMI.2023.3275156}

\end{thebibliography}

\end{document}